\pdfoutput=1

\documentclass[11pt]{article}

\usepackage{acl}

\usepackage{times}
\usepackage{latexsym}

\usepackage[T1]{fontenc}

\usepackage[utf8]{inputenc}

\usepackage{microtype}

\usepackage[usestackEOL]{stackengine}
\usepackage{multirow}

\title{Speech Detection Task Against Asian Hate:\\ BERT the Central, While Data-Centric Studies the Crucial}

\author{Xin Lian \\
  Jinan University - University of Birmingham Joint Institute \\
  \texttt{xxl852@alumni.bham.ac.uk}\\
  \texttt{xinthelian@hotmail.com}}

\begin{document}
\maketitle
\strutlongstacks{T}
\begin{abstract}

With the COVID-19 pandemic continuing, hatred against Asians is intensifying in countries outside Asia, especially among the Chinese. There is an urgent need to detect and prevent hate speech towards Asians effectively. In this work, we first create \textit{COVID-HATE-2022}, an annotated dataset including 2,025 annotated tweets fetched in early February 2022, which are labeled based on specific criteria, and we present the comprehensive collection of scenarios of hate and non-hate tweets in the dataset. Second, we fine-tune the BERT model based on the relevant datasets and demonstrate several strategies related to the “cleaning” of the tweets. Third, we investigate the performance of advanced fine-tuning strategies with various model-centric and data-centric approaches, and we show that both strategies generally improve the performance, while data-centric ones outperform the others, and it demonstrates the feasibility and effectiveness of the data-centric approaches in the associated tasks.
\end{abstract}

\section{Introduction}

In early 2020, the first confirmed case of COVID-19 emerged in Wuhan, China, and after just a few months, the COVID-19 outbreak swept the world with incredible speed and devastation. This pandemic has now had a lasting impact on the world’s economy, society, and politics, while individuals have suffered from epidemic’s economic, physical, and mental health challenges.

In the wake of the epidemic, anti-Asian hatred is intensifying among non-Asian countries, especially among the Chinese. In the US, COVID-19 has enabled the spread of racism and created national insecurity, fear of foreigners, and general xenophobia, which may be associated with the increase in anti-Asian hate crime during the pandemic \citep{vachuska2020initial}. A study suggests former President Donald Trump's inflammatory rhetoric around the coronavirus, which is believed to have originated in China, helped spark anti-Asian Twitter content and “likely perpetuated racist attitudes”: In the week after former President Donald Trump tweeted about the “Chinese Virus”, the number of coronavirus-related tweets with anti-Asian hashtags rose precipitously \citep{hswen2021association}. The US has experienced a rash of violent attacks on people of Asian descent, and naming a disease after a place or a group of people is stigmatizing, according to public health experts \citep{hswen2021association}.

Politicians’ intentional attitudes may be the direct cause, but not the only reason for the increase in Asian hate crimes. There is no effective intervention by social media against related hate speech, including hate speech detection and removal tasks: As algorithms gather user engagement with this toxic content, social media platforms recommend increasingly extreme content to users until their feeds become fully dominated by only the most extreme content, possibly leading to hate crimes \citep{reja2021trump}. Unfortunately, the news media, driven by political interests, has not contributed any to the elimination of Asian hate. The way that the media is covering and the way that people are understanding anti-Asian hate at this moment, in some ways, draws attention to these long-standing anti-Asian biases in U.S. society \citep{yam2021viral}. There is no time to lose in the fight against Asian hate. It’s time to detect and clean up the hate speech against Asians on social media, including Twitter.

\textbf{Related Work.} As social media grows and becomes more influential in the information age, social media are responsible for detecting and preventing hate speech effectively. However, due to the large amount of information, manual detection cannot meet the expectations \citep{li2021covid}. There are various studies on hate speech detections, based on linguistic methods, n-grams, word2vec \citep{nobata2016abusive}, and logistic regression \citep{ousidhoum2019multilingual}, Convolutional Neural Networks, Explainable AI \citep{hardage2020hate}, and Network methods \citep{fan2020stigmatization}. With the advent of Transformers \citep{vaswani2017attention} structure and transfer learning in AI, BERT \citep{devlin2018bert} becomes one of the most popular models for hate speech detection, and an increasing number of studies have shown the dominant performance of BERT in terms of detection tasks (\citealp{liu2019nuli}; \citealp{vishwamitra2020analyzing}; \citealp{li2021covid}; \citealp{rajput2021hate}; \citealp{caselli2020hatebert}; \citealp{saleh2021detection}; \citealp{mohtaj2022feature}), or in general, text classification task. In particular as the number of hate crimes against Asians skyrocketed during the COVID-19 pandemic, more studies of hate speech towards Asians and its detection appeared (\citealp{vidgen2020detecting}; \citealp{alshalan2020detection}; \citealp{tahmasbi2021go}; \citealp{he2021racism}). Most relevant studies focus on how to improve the model itself (fine-tuning and its techniques, or performance between multiple models), or how hate speech datasets look and how to interpret the data, while few have attempted to assess both the data and the model to create a more targeted or custom model, or we say, data-centric studies, instead of focusing on the model and the data separately. In addition to some data-centric studies about how to make the task explainable \citep{mathew2020hatexplain}, we need more data-centric research – we need to understand both the hate speech against Asians and how the data interacts with the model to gain better performance in specific tasks.

\textbf{Our Contributions}. In this paper, we make the following key contributions:

\begin{itemize}
    \item An annotated dataset, \textit{COVID-HATE-2022}, is presented. \textit{COVID-HATE-2022} includes 2,035 annotated tweets fetched in early February 2022, which are labeled based on specific criteria (so the tweets are labeled consistently). Further, we present the typical and comprehensive scenarios for hate tweets towards Asians in our dataset.
    \item We fine-tune BERT model with dataset \textit{COVID-HATE-2022} and other relevant datasets, and evaluate the corresponding models. We further test if the following strategies are effective: fine-tune the models with the respective ``clean''\footnote{To see how the tweets are preprocessed to be “clean”, see Section~\ref{experimentsoverview}}  dataset, and train the model with the original dataset while validating with the respective “clean” one (we also consider the opposite). Our results show that this strategy is not effective when fine-tuning.
    \item We conduct both model-centric and data-centric fine-tuning techniques. We first perform model-centric fine-tuning techniques, including discriminate fine-tuning, gradual unfreezing, and implementing warmup steps, on the respective fine-tuned BERT models. Then we deploy the data by trimming and augmenting referring to our error analysis. Based on our results, advanced fine-tuning techniques are generally effective, and data-centric methods outperform the rest. The result demonstrates the feasibility and effectiveness of data-centric approaches.
\end{itemize}



\section{Annotated Data} \label{annotateddata}

For the annotated dataset COVID-HATE-2022, we fetched the tweets from Feb 1, 2022 to Feb 13, 2022 with Twitter API and tweepy , based on the keywords part of which was introduced by \citet{he2021racism}:

\begin{itemize}
    \item \textbf{COVID-19:} \textit{coronavirus, covid-19, covid19, corona virus}
    \item \textbf{Hate keywords towards the Chinese:} \textit{\# CCPVirus, \#ChinaDidThis, \#ChinaLiedPeopleDied, \#ChinaVirus, \#ChineseVirus, Chinese virus, \#ChineseBioterrorism, \#FuckChina, \#KungFlu, \#MakeChinaPay, \#wuhanflu, \#wuhanvirus, wuhan virus, chink, chinky, chocky, cina, communistvirus, cokin}
    \item \textbf{Hate keywords towards Asian countries other than China, or towards Asian countries in general:} \textit{churka, ting tong, coolie, dink, slant, slant eye, slopehead, yokel}
\end{itemize}

One of the limitations of keyword-based text extracting is that words can be obfuscated in different ways, in the case of automatic content moderation \citep{nobata2016abusive}, or as a consequence of the use of social media for communication (consider, for example, the tendency in some posts to replace letters with similar-looking numbers, e.g., “E”s with “3”s, “l”s with “1”s, “o”s with “0”s, and so on \citep{kovacs2021challenges}.) As a result, a wider number of keywords are included in our tweet fetching:

\begin{quote}
\textit{ch1nk, ch1nky, ch0nky, c1na, c0kin, s1ant, y0kel}
\end{quote}

Further, we add more keywords that are not mentioned in the previous works and also appear in a certain number of relevant hate tweets:

\begin{quote}
\textit{ching chong, chingchong, ch1ng chong, ch1ng ch0ng, chicom, ch1com}
\end{quote}

The most crucial and difficult task is how to define ``hate speech''. There is no universally accepted definition of hate speech \citep{nobata2016abusive}, and one may consider a text as hate speech as soon as he or she with a certain ethnic background feels offensive about it. In this paper, we provide a general criterion of hate tweet towards the Asians:

\begin{quote}
\textit{``Hate Speech'' towards Asians should directly or indirectly exacerbate the discrimination and injustice towards the Asians, including the Chinese, in the face of the epidemic.}
\end{quote}

Based on the given criterion, we made a list of typical scenarios of a “hate tweet” based on the study of the dataset. A hate tweet should:

\begin{itemize}
    \item directly shows hatred or racism towards the Asians\footnote{For instance, a “go-back-to-your-country” tweet or a tweet talking slurs to the Asians.}, or
    \item suggests that COVID-19 is originated (or made, designed, unleashed, etc.) in China explicitly or implicitly\footnote{For instance, a tweet may use the relevant terms “China/Wuhan/Chinese/CCP Virus” or “Kung Flu” as they are in normal usage. Unless the user uses the terms for reference or counterspeeches. For more information on what is a counterspeech, see \url{https://counterspeech.fb.com}.}, or
    \item uses offensive terms from a user who is not an Asian\footnote{For instance, a tweet describing an eye with the word “chinky” from a user not being Asian is recognized as a hate tweet.}, or
    \item expresses hatred towards Asians in other cases.
\end{itemize}

Note that, there may be scenarios that a tweet which is not a hate tweet is easily misclassified. A comprehensive collection of typical examples of hate and non-hate tweets for this research is shown in Appendix~\ref{sec:appendix1}, and Table~\ref{tab:covid2022overview} is an overview of the annotated dataset.

\begin{table}
\centering
\begin{tabular}{l|l}
\hline
No. of tweets & 2,035\\\hline
No. of hate tweets & 497\\\hline
No. of non-hate tweets & 1,538\\\hline
Portion of hate and non-hate tweets & 0.24/0.76\\\hline
\end{tabular}
\caption{Overview of the dataset \textit{COVID-HATE-2022}.}
\label{tab:covid2022overview}
\end{table}


\section{Preliminaries and Methodology}

BERT \citep{devlin2018bert}, or Bidirectional Encoder Representations from Transformers, which is a Transformer-based model, has become one of the most popular and outperforming models for hate speech detection tasks, or in general, text classification tasks, and it is proved to be the most effective performing model in a similar subtask \citep{basile2019semeval}. In this paper, the fine-tuned BERT will be used for the hate speech detection task.

BERT-base model can serve sequences with no more than 512 tokens, so it is likely that some rarely long texts would have to be truncated, and there are various tactics to do this \citep{sun2019fine}. Note that, however, every tweet should be no more than 280 characters, so no text is going to be truncated. After data preprocessing, the model is ready for being fed with the data.

In the following experiments, we use metrics including accuracy score and Matthews Correlation Coefficient (\textbf{MCC}) \citep{matthews1975comparison}, and the latter is shown to be much more effective than the mainstream choices, including accuracy score and F1 score, in a binary classification evaluation \citep{chicco2020advantages}.

\begin{table}[htbp]
  \centering
  \small
    \begin{tabular}{|c|c|c|}
    \hline
    \multicolumn{1}{|r|}{} & Predicted Positive & Predicted Negative \\    
    \hline
    \Longstack{Actual\\ Positive} & \Longstack{True Positive\\(TP)} & \Longstack{False Negative\\(NF)} \\
    \hline
    \Longstack{Actual\\ Negative} & \Longstack{False Positive\\ (FP)} & \Longstack{True Negative \\(TN)}\\
    \hline
    \end{tabular}%
    \caption{The confusion matrix.}
  \label{tab:confusion}%
\end{table}%

The accuracy score is defined as the ratio between the number of correctly classified samples and the overall number of samples. However, it is not a reliable metric when the dataset is unbalanced, i.e., when the number of samples in one class is much larger (or smaller) than the number of samples in the other classes since it provides an over-optimistic estimation of the classifier ability on the majority class (\citealp{sokolova2006beyond}; \citealp{chicco2020advantages}). MCC is an effective solution to redress class imbalances, and it is a contingency matrix method of calculating the Pearson product-moment correlation coefficient (Powers, 2020) between actual and predicted values. Given the confusion matrix (shown in Table~\ref{tab:confusion}), MCC is given by

\[{\scriptstyle
MCC=\frac{TP\cdot TN-FP\cdot FN}{\sqrt{(TP+FP)\cdot(TP+FN)\cdot(TN+FP)\cdot(TN+FN)}}\in[-1,1]}
\]

$MCC=-1$ stands for complete misclassification, and $MCC=1$ stands for exact classification, while $MCC=0$ is no better than random prediction.

\section{Experiments}
\subsection{Datasets Overview}

The following datasets are used to fine-tune and test models:

\textbf{\textit{COVID-HATE}\footnote{For code and data of the research, see \url{http://claws.cc.gatech.edu/covid}}, \textit{COVID-HATE-2022}, and \textit{COVID-HATE-CON}.} \textit{COVID-HATE} \citep{he2021racism} is the largest dataset of anti-Asian and counterspeech spanning 14 months (from Jan 15, 2020 to Mar 26, 2021), containing over 206 million tweets and a social network with over 127 million nodes. 3,555 tweets are annotated to train the classifier model. The tweets were fetched based on the keyword list with categories COVID-19, Hate Speech, and Counterspeech. In this paper, we set all tweets that belong to the hate-speech class. \textit{COVID-HATE-2022}, whose keyword list comprises part of ours, can be considered as a supplement to \textit{COVID-HATE} in response to tweets in early 2022. Consequently, \textit{COVID-HATE-CON}, the concatenation of \textit{COVID-HATE} and \textit{COVID-HATE-2022}, is indeed an extended \textit{COVID-HATE}.

\textbf{\textit{HatEval}}\footnote{To get access to the data of the research, visit \url{http://hatespeech.di.unito.it/hateval.html}}. The complete dataset \textit{HatEval} includes English and Spanish splits, and we only consider the English split here, which contains 18,000\footnote{The dataset is split into test and training sets, and 18,000 is the number of texts in both sets. The information for \textit{HatEval} in Table~\ref{tab:overviewdata} is also for both sets as well.} tweets annotated for hate speech against migrants and women. In their annotation process, four classes are considered: hate speech, target range, aggressiveness, and others. In our investigation, we assigned tweets with positive hate-speech labels to the hate-speech category, while others belong to the non-hate-speech category. We will retrain the BERT model based on its training set, and evaluate the fine-tuned model based on Asian-hate data to see if the models detecting different fields of hate speech are “compatible”.

\begin{table*}[htbp!]
\centering
\small
\begin{tabular}{ccccc}
\hline
\textbf{Data}  & \multicolumn{1}{p{8em}}{\textbf{No. of Hate Tweets}} & \multicolumn{1}{p{8em}}{\textbf{No. of Non-Hate Tweets}} & \multicolumn{1}{p{12em}}{\textbf{Portion of Hate/Non-Hate Tweets}} & \multicolumn{1}{p{7em}}{\textbf{No. of Samples}} \\
    \hline\hline
    \textit{COVID-HATE-2022} & 497   & 1,538 & 0.24/0.76 & 2,035 \\\hline
    \textit{COVID-HATE} & 429   & 1,861 & 0.19/0.81 & 2,290 \\\hline
    \textit{COVID-HATE-CON} & 926   & 3,399 & 0.21/0.79 & 4,325 \\\hline
    \textit{HatEval} & 7,566 & 10,434 & 0.42/0.58 & 18,000 \\\hline
\end{tabular}
\caption{An overview of the datasets in the experiments}
\label{tab:overviewdata}
\end{table*}

Table~\ref{tab:overviewdata} is a summary of the basic information of the datasets that are planned to be used in the following experiments. Except for \textit{HatEval}, all the datasets are imbalanced, which means in the later experiments, a relatively high accuracy based on the respective datasets may not indicate high performance. The models in the original research associated with \textit{COVID-HATE} and \textit{HatEval} datasets are not evaluated by MCC, so this paper will also see if the models are indeed effective.

\subsection{Experiments Overview} \label{experimentsoverview}

In the following experiments, we are aiming to investigate:

\begin{itemize}
    \item the performance of fine-tuned BERT models on Asian hate speech detection tasks trained with different datasets,
    \item the performance of fine-tuned BERT models on hate speech detection tasks trained with different datasets, whose tweets are “clean”, i.e., preprocessed by dropping the usernames being @, the hashtags, URLs, and emojis in every tweet\footnote{This is realized by the preprocessing library for tweet data in Python. Visit \url{https://pypi.org/project/tweet-preprocessor}.}.
    \item the performance of fine-tuned BERT models on hate speech detection tasks trained with advanced fine-tuning techniques, which are model-centric, as well as the models with data-centric studies.
\end{itemize}

\subsection{Experiments Setup}

In our implementation, we use the tokenizers and fine-tune the BERT model based on the pre-trained \texttt{BERT-base-uncased} model pipeline via the \texttt{transformer} library, Hugging Face\footnote{The model is available at \url{https://github.com/huggingface/transformers}} \citep{wolf2019huggingface}.

During data preprocessing, we set the batch size to 24. Suppose after tokenization, the largest number of tokens of sequences in batch $i$ is $L_i$, then the maximum length of sequences in batch $i$ (i.e., the length that all the sequences in batch $i$ should be padded to) is given by:
\[maxlength_i=\lceil\frac{L_i}{128}\rceil\times128\]

and all the padding should be added to the end of each original sequence of tokens. Each dataset will be randomly divided into the following proportions:
\[train:validate:set=8.1:0.9:1\]

The original and “clean” Twitter data will be included in all datasets.

The \textit{AdamW} algorithm \citep{loshchilov2017decoupled} will be used as an optimization during training with a learning rate of \texttt{2e-5}, and no warmup steps will be taken. Each model will be trained with 4 epochs and both training validation processes will be processed in each epoch.

In the following experiments:

\begin{itemize}
    \item The testing results only with positive MCC coefficients are displayed, which will be referred to as ``valid'' results in the followings,
    \item the respective accuracy score will be in bold if it exceeds the portion of non-hate tweets in the respective dataset, and
    \item the MCC score will be in bold if it is the highest among others in its respective section.
\end{itemize}

\subsection{Based Model Fine-tuning Experiments}

\begin{table*}[htbp]
\centering
\small
\begin{tabular}{p{15em}ccc}
\hline
\textbf{Training and Validation Set from} &	\textbf{Test Set from} &	\textbf{Accuracy} &	\textbf{MCC}\\
\hline\hline
\textit{COVID-HATE-2022}	& / & / & / \\\hline
\multirow{3}{10em}{\textit{COVID-HATE}} & \textit{COVID-HATE-2022} & \textbf{0.779412} &	0.13\\
& \textit{COVID-HATE-CON} &	\textbf{0.794457} &	\textbf{0.185}\\
& \textit{HatEval}	& \textbf{0.580889} &	0.032\\\hline
\textit{COVID-HATE-CON}	& /	& / & /\\\hline
\textit{HatEval}	& / & / & /\\\hline
\textit{COVID-HATE-2022} (clean)	& / & / & /\\\hline
\multirow{2}{6em}{\textit{COVID-HATE} (clean)} & \textit{COVID-HATE-CON} &	\textbf{0.792148} &	0.16\\
& \textit{HatEval} & \textbf{0.580889} &	0.032\\\hline
\textit{COVID-HATE-CON} (clean)	& /	& /	&/\\\hline
\textit{HatEval} (clean)	& /	& /	&/\\\hline
\end{tabular}
\caption{Valid testing results of based fine-tuned models.}
\label{tab:task1}
\end{table*}

Table~\ref{tab:task1} lists the valid testing results of the relevant fine-tuned models. The BERT model fine-tuned with \textit{COVID-HATE} has the greatest performance given the test set of \textit{COVID-HATE-CON}, whose MCC coefficient reaches 0.185, exceeding most other cases with zero (or even negative) MCC coefficients, and it demonstrates the BERT model fine-tuned with \textit{COVID-HATE} is more compatible with hate speech detection tasks given tweets data with an extended time. Further, all other fine-tuned BERTs show only marginal performance when dealing with \textit{COVID-HATE-2022} data.

BERTs fine-tuned with “clean” data failed to deal with the detection task\footnote{All cases with “clean” data being the test set are not displayed here, which means their results are not even valid according to our criteria.}, and they appear to be less effective in detecting "clean" tweets. One reliable speculation for this result is because much relevant information of the tweets, especially for the hate tweets, is included as hashtags\footnote{Numerous examples of “relevant information is included as hashtags” are available in the Appendix~\ref{sec:appendix1}.}.

\subsection{Model-Centric Experiments: Advanced Fine-tuning Techniques\footnote{In the rest of this section, we would not fine-tune any BERT model with \textit{HatEval}.}}

In spite of the fact that some of the fine-tuned models are effective at detecting hate speech, the results are not as good as expected. In this section, we will try to figure out whether the models are improved by advanced fine-tuning techniques, which are raised to deal with few-sample scenarios or catastrophic forgetting issues \citep{howard2018universal}.

\textbf{Discriminative Fine-tuning.} The idea is based on the fact that different layers capture different types of information \citep{yosinski2014transferable}. Then instead of using the same learning rate for all layers of the models, one can tune each layer with different learning rates by discriminative fine-tuning, and more specifically, the scheme is accomplished by setting the learning rate of the top layer and using a multiplicative decay rate to decrease the learning rate layer-by-layer from top to bottom\footnote{An alternative name for this technique is called Layer-wise Learning Rate Decay (LLRD) \citep{zhang2020revisiting}.} \citep{howard2018universal}. In the corresponding experiments, we follow the scheme of layer-wise learning rate decay, set the learning rate of the initial layer to \texttt{1.9e-5} with a decay rate of 0.9 after each layer, then reset the learning rate of the last layer (pooling and regression layer) to \texttt{2e-5}.

\textbf{Freezing Layers, Gradual Unfreezing.} In an original fine-tuning task, the parameters of all layers are trained at once. This could be computationally costly when dealing with large models with millions of parameters, and it may risk catastrophic forgetting. In a more advanced way, the concept of gradual unfreezing is proposed - first unfreeze the last layer and fine-tune all other layers for one epoch, then unfreeze the next lower frozen layer and repeat, until all layers are fine-tuned until convergence at the last iteration (\citealp{howard2018universal}; \citealp{yosinski2014transferable}). In the corresponding experiments, we will evaluate the performance of fine-tuned models with gradual unfreezing of the last 4, 8, 12 layers (with corresponding 4, 8, and 12 epochs).

\textbf{Warmup Steps, Slanted Triangular Learning Rates.} Setting warmup steps is a tactic raised to avoid the expected gradients of the parameters near the output layer of the Transformer being large, namely, it helps the model quickly converge to a suitable region of the parameter space at the beginning of training and then refine its parameters \citep{howard2018universal}. Further, the concept of Slanted Triangular Learning Rates \citep{howard2018universal} is proposed, which additionally sets the rules for the cutting fraction for steps to warmup. It is worth noting that some research claims that the warmup stage can be safely removed \citep{xiong2020layer}. In the corresponding experiments, we will investigate the performance of models with 25, 50, 75, and 100 warmup steps.

\begin{table*}[h!]
\centering
\scriptsize
{\renewcommand{\arraystretch}{1.1}
\begin{tabular}{p{13em}|p{17em}p{10em}p{4em}p{2em}}
\hline
\textbf{Training and Valid Sets from} &	\textbf{Fine-tuning Technique} &	\textbf{Testing Set from} & \textbf{Accuracy}	& \textbf{MCC}\\
\hline\hline
\multirow{11}{13em}{\textit{COVID-HATE-2022}} &	None & / & / & /\\\cline{2-5}
	&Discrimination	& \textit{HatEval} &	0.580000 &	0.015\\\cline{2-5}
	&25 Warmup Steps	& \textit{HatEval} &	0.579778 &	0.012\\\cline{2-5}
	&50 Warmup Steps & / & / & / \\\cline{2-5}
	&75 Warmup Steps & / & / & / \\\cline{2-5}
	&\multirow{2}{10em}{100 Warmup Steps} &	\textit{COVID-HATE-CON} & 0.785219 &	0.082\\
		&&\textit{HatEval} &	\textbf{0.581444}&	0.035\\\cline{2-5}
	&Gradually Unfreeze the Last 4 Layers& / & / & / \\\cline{2-5}
	&\multirow{2}{10em}{Gradually Unfreeze the Last 8 Layers}&	\textit{COVID-HATE-CON} &	0.789838 &	\textbf{0.126}\\
		&&\textit{HatEval}&	\textbf{0.582667}&	0.045\\\cline{2-5}
	&Gradually Unfreeze the Last 12 Layers	& / & / & / \\\cline{2-5}
	\hline
\multirow{15}{13em}{\textit{COVID-HATE}}	& \multirow{3}{*}{None} &	\textit{COVID-HATE-2022} &	\textbf{0.779412} &	0.130\\
		&& \textit{COVID-HATE-CON}&	\textbf{0.794457}&	\textbf{0.185}\\
		&&\textit{HatEval} &	\textbf{0.580889}&	0.032\\\cline{2-5}
	&\multirow{2}{*}{Discrimination}	&\textit{COVID-HATE-CON}&	0.780600&	0.008\\
		&&\textit{HatEval}&	\textbf{0.580556}&	0.030\\\cline{2-5}
	&25 Warmup Steps	&\textit{HatEval}&	0.579667&	0.012\\\cline{2-5}
	& 50 Warmup Steps &	\textit{COVID-HATE-CON} &	0.750577&	0.029\\\cline{2-5}
	&\multirow{2}{*}{75 Warmup Steps}&	\textit{COVID-HATE-CON}&	0.771363&	0.014\\
		&&\textit{HatEval}&	0.577222&	0.014\\\cline{2-5}
	&\multirow{2}{*}{100 Warmup Steps}&	\textit{COVID-HATE-CON}&	\textbf{0.792148}	&0.160\\
		&&\textit{HatEval}&	\textbf{0.580889}&	0.032\\\cline{2-5}
	&Gradually Unfreeze the Last 4 Layers	&/&/&/\\\cline{2-5}
	&Gradually Unfreeze the Last 8 Layers	&/&/&/\\\cline{2-5}
	&\multirow{2}{10em}{Gradually Unfreeze the Last 12 Layers}	&\textit{COVID-HATE-CON}&	0.787529&	0.111\\
		&&\textit{HatEval}	&0.580333&	0.028\\\cline{2-5}
		\hline
\multirow{13}{13em}{\textit{COVID-HATE-CON}}&	None	&/&/&/\\\cline{2-5}
	&Discrimination&	\textit{HatEval}&	0.579889&	0.012\\\cline{2-5}
	&25 Warmup Steps&/&/&/\\\cline{2-5}
	&\multirow{2}{*}{50 Warmup Steps}&	\textit{COVID-HATE-2022}&	\textbf{0.779412}&	\textbf{0.130}\\
		&&\textit{COVID-HATE-CON}&	0.787529&	0.092\\\cline{2-5}
	&75 Warmup Steps &/&/&/\\\cline{2-5}
	&\multirow{2}{*}{100 Warmup Steps}&	\textit{COVID-HATE-CON}&	0.789838&	0.126\\
		&&\textit{HatEval}&	0.579556	&0.009\\\cline{2-5}
	&Gradually Unfreeze the Last 4 Layers&/&/&/\\\cline{2-5}
	&\multirow{3}{10em}{Gradually Unfreeze the Last 8 Layers}&	\textit{COVID-HATE-2022}&	\textbf{0.769608}&	0.032\\
		&&\textit{COVID-HATE-CON}&	0.785219&	0.107\\
		&&\textit{HatEval}&0.579556	&0.030\\\cline{2-5}
	&Gradually Unfreeze the Last 12 Layers&/&/&/\\\cline{2-5}
\hline
\end{tabular}
}
\caption{Valid testing results of experiments with models and advanced fine-tuning techniques.}
\label{tab:task2}
\end{table*}

\textbf{Experiments Results and Analysis.}    Table~\ref{tab:task2} shows the valid test results of the experiments on advanced fine-tuning techniques. The advanced techniques for fine-tuning are generally more beneficial to the performance of the models: BERT fine-tuned with \textit{COVID-HATE-2022} reaches the best-performing score when fine-tuned with the last 8 layers being gradually unfrozen, and BERT fine-tuned with \textit{COVID-HATE-CON} reaches the best-performing score when fine-tuned with additional 50 warmup steps, both of which perform better than the corresponding vanilla ones. However, BERT fine-tuned with \textit{COVID-HATE} maintains its highest testing result with the vanilla version, which indicates the advanced fine-tuning strategies do not necessarily distribute progress to the models. From a general perspective for these datasets, 100 warmup steps and the last 8 to 10 layers for gradual unfreezing are relatively good choices when considering advanced fine-tuning strategies.

\subsection{Data-Centric Experiments: Improve Our Data}

Your model architecture is good enough: In recent publications, 99\% of the papers were model-centric with only 1\% being data-centric, and we assume they did their homework right. Data is the food for AI, and one may need to source and prepare high-quality ingredients to “feed” the model \citep{ng2021mlops}. Data-centric approaches have been proven to be effective, sometimes even better than model-centric ones for particular cases, especially for computer vision tasks \citep{ng2021mlops}. In this subsection, we do analysis based on the data \textit{COVID-HATE-2022} itself.

\textbf{Error Analysis and Data Deploying.} From previous results, we can see the BERT fine-tuned with \textit{COVID-HATE-2022} performs less well than expected – it is no better than random classification because the corresponding MCC score is 0 when testing with the test set from \textit{COVID-HATE-2022}. We notice that all the error predictions are Type II errors, namely, all the test samples being misclassified are labeled as hate speeches, while the model classifies them to non-hate speeches. Based on the facts, we may deploy our data \textit{COVID-HATE-2022} based on the strategies:

\begin{itemize}
    \item \textbf{Augment your data.} We can include multiple duplicates of hate speeches into the training set. We wish the model could learn multiple times about “how to detect hate speech”.
    \item \textbf{Trim your data.} We can raise the percentage of hate speeches in the training set and remove the "perturbed" ones\footnote{For more details, see Appendix~\ref{sec:appendix2}.} by trimming some of the non-hate speech. One potential purpose of doing this is to mitigate the negative effects of perturbed instances.
\end{itemize}

\begin{table*}[htbp!]
\centering
\small
\begin{tabular}{p{15em}p{6em}p{6em}p{10em}p{4em}}
\hline
\textbf{Data} &	\textbf{No. of Hate Tweets}&	\textbf{No. of Non-Hate Tweets} &	\textbf{Portion of Hate/Non-Hate Tweets} &	\textbf{No. of Samples}\\\hline\hline
\textit{COVID-HATE-2022}	&497&	1,538&	0.24/0.76&	2,035\\\hline
\textit{COVID-HATE-2022-TRIM}	&415	&437&	0.49/0.51&	852\\\hline
\textit{COVID-HATE-2022-AGU}&	994	&1,538&	0.39/0.61&	2,532\\\hline
\end{tabular}
\caption{Overview of the deployed datasets in the data-centric experiments.}
\label{tab:deployedoverview}
\end{table*}

For the rest of this subsection, we will consider the deployed datasets from \textit{COVID-HATE-2022}: \textbf{\textit{COVID-HATE-2022-TRIM}}, some of whose non-hate speech data are trimmed because they are irrelevant to any key themes of our research, and \textbf{\textit{COVID-HATE-2022-AGU}}, which is indeed \textit{COVID-HATE-2022} with its hate speech data being duplicated. Table~\ref{tab:deployedoverview} is an overview of the datasets for fine-tuning BERT in the following.

\begin{table*}[h!]
\centering
\footnotesize
{\renewcommand{\arraystretch}{1.1}
\begin{tabular}{p{22em}p{12em}p{5em}p{3em}}
\hline
\textbf{Training and Valid Sets from}&	\textbf{Testing Set from} &	\textbf{Accuracy}&	\textbf{MCC}\\\hline\hline
\textit{COVID-HATE-2022} &/&/&/\\\hline
\multirow{2}{22em}{\textit{COVID-HATE-2022} (Gradually Unfreeze the Last 8 Layers)}&	\textit{COVID-HATE-CON}&	0.789838&	\textbf{0.126}\\
	&\textit{HatEval}&	\textbf{0.582667}&	0.045\\\hline
\multirow{3}{22em}{\textit{COVID-HATE-2022-TRIM}}&	\textit{COVID-HATE-2022}&	0.284314&	\textbf{0.135}\\
	&\textit{COVID-HATE-CON}&	0.265589&	0.085\\
	&\textit{HatEval}&	0.444444&	0.051\\\hline
\multirow{3}{22em}{\textit{COVID-HATE-2022-TRIM} (with number of epochs 5)}	&\textit{COVID-HATE-2022}&	0.784314&	\textbf{0.184}\\
	&\textit{COVID-HATE-CON}&	0.782910&	0.094\\
	&\textit{HatEval}&	0.579222&	0.034\\\hline
\multirow{3}{22em}{\textit{COVID-HATE-2022-AGU}}&	\textit{COVID-HATE-2022}&	\textbf{0.779412}&	\textbf{0.130}\\
	&\textit{COVID-HATE-CON}&	0.780600&	0.034\\
	&\textit{HatEval}&	0.579222&	0.017\\\hline
\multirow{4}{22em}{\textit{COVID-HATE-2022-AGU}
(with number of epochs 5)}&	\textit{COVID-HATE-2022}&	\textbf{0.774510}&	0.065\\
	&\textit{COVID-HATE-CON}&	\textbf{0.787529}&	\textbf{0.111}\\
	&\textit{HatEval}&	0.578889&	0.011\\
	&\textit{COVID-HATE-2022-AGU}&	0.606299&	0.034\\\hline
\end{tabular}
}
\caption{Valid testing results of the experiments with data-centric models.}
\label{tab:task3}
\end{table*}

\textbf{Experiments Results and Analysis.}    Table~\ref{tab:task3}\footnote{We include the result of BERT model derived from \textit{COVID-HATE-2022} (Gradually Unfreeze the Last 8 Layers, the best result among the model-centric experiments) here to compare with the results of the data-centric ones.} shows the valid testing results of the relevant experiments in this subsection. By trimming the dataset \textit{COVID-HATE-2022} (and deriving the dataset \textit{COVID-HATE-2022-TRIM}) and increasing one epoch during the training process, the corresponding fine-tuned model performs the most efficiently when dealing with the hate speech detection upon \textit{COVID-HATE-2022}, and generally, the model outperforms the vanilla one and the ones trained with the advanced fine-tuning techniques. Additionally, by augmenting \textit{COVID-HATE-2022} (and deriving the dataset \textit{COVID-HATE-2022-AUG}) with its hate-speech texts, the corresponding fine-tuned model also outperforms the vanilla one and the ones trained with the advanced fine-tuning techniques in general. Therefore, in this Asian-hate-speech-detection case, data-centric approaches are more effective than model-centric approaches.

\section{Conclusions and Discussions}

In this work, we notably present a comprehensive analysis of the forms and criteria of Asian-hate tweets, and our work paves the way toward anti-Asian-hate messaging campaigns as a potential solution to hate speech on social media. Moreover, our study supports the predominance of data-centric methods when dealing with the particular hate speech detection task: although we experiment with both model-centric and data-centric approaches, and both achieve better results, our work sheds light on future data-centric research in the relevant area.

Our work does have a few limitations. Compared to other major hate speech datasets, \textit{COVID-HATE-2022} includes tweets within a short time span, so the information absorbed by the derived model is much more focused, and therefore, may not be robust in further detection tasks. Moreover, while both model-centric and data-centric advanced fine-tuning strategies increase performance, the results are lower than anticipated - all MCC scores are less than 0.2, and test scores are sensitive even with slightly deployed data. The models may achieve a higher MCC score as they perform in a larger corpus of Asian hate. The truth is, the Transformer-based classifier models may be more sensitive than we expected.

In the section~\ref{annotateddata}, we present how the text is complicated and perturbed. According to research about the effect of adversarial behavior and augmentation for cyberbullying detection (or in general, toxic comments) regarding small corpora, the less training data is available, the more vulnerable models might become \citep{emmery2022cyberbullying}. Furthermore, it is demonstrated that model-agnostic lexical substitutions, or in general, perturbations, significantly hurt classifier performance, and when the perturbed samples are used for augmentation, models become robust against word-level perturbations at a slight trade-off in overall task performance \citep{emmery2022cyberbullying}. Therefore, data trimming and data augmentation strategies could be used to achieve the right balance between performance and robustness. Based on the study, future work may concentrate not only on data-centric research but also on the trade-off between performance and robustness considering classification based on small corpora to mitigate the effect of adversarial behavior.

\bibliography{anthology,custom}
\bibliographystyle{acl_natbib}

\appendix

\section{Hate and Non-Hate Tweets Examples}
\label{sec:appendix1}

Here is a collection of typical examples of hate and non-hate tweets.

\textbf{Disclaimer}: The appendix contains material that many will find offensive or hateful, especially for Asians; however, this cannot be avoided owing to the nature of the work.

\subsection{Hate Tweets Examples}
\begin{itemize}
    \item The “go-back-to-your-country” tweet
    \begin{quote}
        \textit{@user @user @user \textbf{Go away and take COVID with you}!!! \#chinaliedpeopledied}
    \end{quote}
    \begin{quote}
        \textit{@user Koreans are literally living rent free of all \textbf{ch1nk}’s head. \textbf{Get outta twitter}}
    \end{quote}
    \item Explicit shame words (may consider the case of forbidding automatic correction)
    \begin{quote}
        \textit{Deserved. No way should these games be taking place in a \textbf{Pariah State} \#CCP \#chinaliedpeopledied \#WinterOlympics}
    \end{quote}
    \begin{quote}
        \textit{@user @user @user Dimwit \textbf{IT coolie}.}
    \end{quote}    
    \begin{quote}
        \textit{LOVED how USA "subjagated" \textbf{PUNK LOSER WHORE} \#chinaliedpeopledied in ice hockey, shutting them out 8-0}
    \end{quote}
    \item Claiming China is responsible for the pandemic
    \begin{quote}
        \textit{\#chinaliedpeopledied \textbf{Hold CCP accountable} for spreading Wuhan virus and causing global pandemic. No evil deeds shall go unpunished.}
    \end{quote}  
    \begin{quote}
        \textit{"Every lie we tell incurs a debt to the truth. Sooner or later that \textbf{debt is paid}". \#CCPChina \#chinaliedpeopledied}
    \end{quote}
    \item Directly claiming COVID-19 is originated/designed in China/Wuhan Lab
    \begin{quote}
        \textit{I,will never forgive the \textbf{Chinese for unleashing this virus} on us in an act of war. Make no mistake it was an act of war. F**k China. I will never forgive our President and his idiot followers (Dimslee,Newsom et al) for the harm they have done to our seniors and children.}
    \end{quote} 
    \begin{quote}
        \textit{@user Bc \#China owns everything now. \textbf{They also made the \#ChinaVirus}}
    \end{quote}
    \item Implicitly claiming COVID-19 is originated in China by using the offensive terms
    \begin{quote}
        \textit{@user So ridiculous the vaccine does not keep you from getting the \textbf{wuhan 19 virus} or spreading it}
    \end{quote}    
    \begin{quote}
        \textit{@user Get well soon from \textbf{\#ChineseVirus!}}
    \end{quote}  
    \item Others
    \begin{quote}
        \textit{The \#CCP, \#Democrat produced \#ChinaVirus \#Plandemic to remove \#PresidentTrump is over. How do you know its over? Time to \#Vote thats how, \#Yahoo pivots. So @user are in a pants pissing panic over their 100\% guaranteed loss of power. \#WakeUpAmerica, it’s \#VaccineTerrorism}
    \end{quote}
    The user believes that Trump's downfall was caused by China, and the tweet may incite the pro-Trump crowd to hate China.
\end{itemize}

\subsection{Non-Hate Tweets Examples}
\begin{itemize}
    \item Counterspeech
    \begin{quote}
        \textit{People who say the virus comes from China, I think you are really stupid. Where did you hear the virus comes from China? Did you hear that from the marketing number? They will only make rumors. \#ChineseVirus}
    \end{quote}
    \begin{quote}
        \textit{@user @user @user @user @user @user Wtf. Whites and blacks called Jeremy lin a ch1nk when he played games in the states. There's racists everywhere in every country.}
    \end{quote}
    \item Propaganda/conspiracy/rumor rather than a hate tweet
    \begin{quote}
        \textit{I’m secretly a chinese national and covert operator. i leaked the virus. i am now in taiwan plotting against their president.}
    \end{quote}
    \begin{quote}
        \textit{@user Working on a new political intrigue novel. Worldwide environmentalist and leftist in the US, conspire the Chinese communists to create a virus that targets the elderly and obese. They wish to reign in the human population and leave only strong workers …}
    \end{quote}
    \item References, regardless of the original sources are biased or not
    \begin{quote}
        \textit{@user @user @user @user @user @user @user @user "The origin of the SARS-CoV-2 virus that causes COVID-19 remains unclear, but recent revelations reinforce the likelihood that the true source was a lab leak from the Wuhan Institute of Virology." Did you forget to read the very first sentence of your own article?}
    \end{quote}
    \begin{quote}
        \textit{"'Early version of Covid-19' is discovered in Chinese lab, fuelling fears that scientists were studying the virus prior to outbreak" : (url)}
    \end{quote}
    \item Discussion rather than a hate tweet
     \begin{quote}
        \textit{@user @user @user Check deeper on what the Sars/Covid virus was and where it came from. Have you heard of chemical warfare? Stuff has to be researched dude. AND AMERICA started the research. Then the Chinese took it over.}
    \end{quote}
    \begin{quote}
        \textit{Chinese vaccine is showing its impact. Zero cure of Corona. Now perpetrators of Corona virus spread are on Almighty Radar}
    \end{quote}
    \item Opinions based on political rumors
    \begin{quote}
        \textit{@user \#F**kIsrael been killing my people for 75 yrs. \#F**kChina been killing Uyghurs for being Muslims.}
    \end{quote}
    \item Using related tags for other topics which are not related to Asian hates (for example, the tweets are actually condemning Biden, or the Democrats, etc.)
    \begin{quote}
        \textit{\#Meanwhile in \#USA \& \#Canada... \#FreedomConvoy \#TruckersForFreedom \#JustSayNo \#Event201 \#OutbreakAnatomyofaPlague \#CoronaVirus \#Deltacron \#IHU \#Omicron \#Florona \#Covid19 \#Delta \#ChinaLiedPeopleDied Bret Baier: This is a big problem for Justin Trudeau}
    \end{quote}   
    \begin{quote}
        \textit{Show Americans the kits! Do the American People feel that they've been lied to? \#HidenBidenLieden \#SouthernBorderCrises \#LetsGoBrandon \#AfghanistanDeaths \#InfationBiden \#ChinaVirus \#BidenWorstPresidentEver \#MassPsychosis \#StupidSonofaBitchBiden \#TruckConvoy2022 \#BidenCrackPipe}
    \end{quote}   
    \item Tweets using related terms with other meanings instead of offensive ones
    \begin{quote}
        \textit{@user There has to be a \textbf{chink in the armor\footnote{If you say that someone has a "chink in their armor", you mean that they have a small weakness in their character or in their ideas which makes it easy to harm them \citep{collins2014cobuild}.}} of your contract. No way they should keep you if you're not happy.}
    \end{quote}    
    \begin{quote}
        \textit{@user @user Lol it's not possible IMDb is unbiased site if this could be possible than Radhe, sadak2, \textbf{coolie no1}\footnote{This tweet uses the word "coolie" to refer to the 2020 (or 1995 original version of) Indian Hindi-language comedy masala movie “Coolie No.1”.} would not get such poor rating ... I knew kangu fans don't have brain}
    \end{quote}    
    \begin{quote}
        \textit{@user My \textbf{rinky dink}\footnote{“Rinky-dink” means having little importance or influence, or old-fashioned or of poor quality \citep{walter2008cambridge}.} home has gone up over 35\% in less than 2 years since I got into it. It's insane, and the youngest adults are hosed. Something is going to break, either a pricing crash or a revolution.}
    \end{quote}
    \begin{quote}
        \textit{Runs farther routes than slant boy\footnote{This tweet uses “slant” here to describe “Slant Boy”, which is the nickname of Michael Thomas, a player in the NFL. For more information, see  \href{https://larrybrownsports.com/football/michael-thomas-mocked-with-slant-boy-nickname-after-zero-catch-game/572137}{here}.}}
    \end{quote}
    \begin{quote}
        \textit{@user @user Cleta the Slack-Jawed Yokel\footnote{This tweet uses the word "yokel" to refer to the song \textit{Cletus The Slack-Jawed Yokel!}, which is the theme song for a segment revolving around Cletus in the episode 22 \textit{Short Films About Springfield.} of the series \textit{The Simpsons}.}}
    \end{quote}
    \item Other tweets that are not related to any of the topics
    \begin{quote}
        \textit{COVID test available tomorrow at CVS (4100 State Highway 121, Carrollton, TX 75010) at 12:00PM 12:10PM 12:20PM 12:30PM 12:40PM 01:20PM 02:20PM 02:30PM 02:40PM 02:50PM 03:10PM 03:20PM 03:30PM 03:40PM 03:50PM : (url)}
    \end{quote}   
    \begin{quote}
        \textit{@user City girls up by 1 point}\citep{https://doi.org/10.48550/arxiv.2206.02114}
    \end{quote}
\end{itemize}

\section{Trimming Your Data: "Pertubed" Tweets Examples}
\label{sec:appendix2}
Typical examples for the “perturbed” scenario are the tweets being fetched just because the usernames are also considered when fetching tweets with the keywords listed, so although the context of the tweet has nothing to do with Coronavirus, hate speech, counterspeech, or any other relevant topics, they may be fetched by the assoicated API. One way to deploy the data is to delete these tweets from the training set.

\begin{quote}
    \textit{I realize that there is not A Single simple straightforward R tutorial to evaluate the quality and linearity of your proteomic bottom up DIA experimental setup, and this makes me sad. Why such a secrecy? \#DIA \#massspectrometry \#dataanalysis \#proteomics}\\
    ------ a tweet from a user with the term “yokel” included
\end{quote}
\begin{quote}
    \textit{this is very chilling}\\
    ------ a tweet from a user with the term “ch1nk” included
\end{quote}
\begin{quote}
    \textit{The old ladies at the nail salon be so rough shit}\\
    ------ a tweet from a user with the term “chink” included
\end{quote}

\end{document}